\documentclass[a4paper,conference]{IEEEtran}
\usepackage[pdftex]{graphicx}
\graphicspath{figures}

\usepackage{tabularx}
\usepackage{booktabs}
\usepackage{dcolumn}
\usepackage{hyperref}

\usepackage[x11names]{xcolor}

\newif\ifproofread

\newcommand{\highlight}[1]{%
\ifproofread
\textcolor{red}{#1}%
\else
#1%
\fi
}

\usepackage{cite}
\ifCLASSINFOpdf
\else
\fi
\usepackage{amsmath}
\usepackage{amsfonts}
\begin{document}
\title{Transformer Reasoning Network for Image-Text Matching and Retrieval}

\author{\IEEEauthorblockN{Nicola Messina, Fabrizio Falchi, Andrea Esuli, Giuseppe Amato}
\IEEEauthorblockA{\textit{Institute of Information Science and Technologies} \\
\textit{National Research Council}\\
Pisa, Italy \\
\{name.surname\}@isti.cnr.it}
}


%Telephone: (800) 555--1212, Fax: (888) 555--1212}


\maketitle

\begin{abstract}
Image-text matching is an interesting and fascinating task in modern AI research. Despite the evolution of deep-learning-based image and text processing systems, multi-modal matching remains a challenging problem. In this work, we consider the problem of accurate image-text matching for the task of multi-modal large-scale information retrieval. State-of-the-art results in image-text matching are achieved by inter-playing image and text features from the two different processing pipelines, usually using mutual attention mechanisms. However, this invalidates any chance to extract separate visual and textual features needed for later indexing steps in large-scale retrieval systems. In this regard, we introduce the Transformer Encoder Reasoning Network (TERN), an architecture built upon one of the modern relationship-aware self-attentive architectures, the Transformer Encoder (TE).
This architecture is able to separately reason on the two different modalities and to enforce a final common abstract concept space by sharing the weights of the deeper transformer layers.
Thanks to this design, the implemented network is able to produce compact and very rich visual and textual features available for the successive indexing step.
Experiments are conducted on the MS-COCO dataset, and we evaluate the results using a discounted cumulative gain metric with relevance computed exploiting caption similarities, in order to assess possibly non-exact but relevant search results. We demonstrate that on this metric we are able to achieve state-of-the-art results in the image retrieval task. Our code is freely available at \url{https://github.com/mesnico/TERN}.
\end{abstract}


% \fi
\IEEEpeerreviewmaketitle

\proofreadfalse

\section{Introduction}

Recent advances in deep learning research brought to life interesting tasks and applications which include joint processing of data from different domains. Image-text matching is an interesting task that consists in aligning information coming from visual and textual worlds, in order to benefit from the complementary richness of these two very different domains. 

Visuals and texts are two important modalities used by humans to fully understand the real world. While text is already a well-structured description developed by humans in hundreds of years, images are basically nothing but raw matrices of pixels hiding very high-level concepts and structures. If we want to obtain an informative textual description of a visual scene we are required not only to understand what are the salient entities in the image, but we need also to reason about the relationships between the different entities, e.g. "The kid \textit{kicks} the ball".
In this respect, it is necessary not only to perceive objects on their own but also understanding spatial and even abstract relationships linking them together.

This has important implications in many modern AI-powered systems, where perception and reasoning play both important roles. In this work, we concentrate our effort on the cross-modal information retrieval research field, in which we are asked to produce compact yet very informative object descriptions coming from very different domains (visual and textual in this scenario). %This is needed for the downstream indexing procedures for fast and efficient retrieval.

Vision and language matching has been extensively studied \cite{vsepp2018faghri,carrara2018pictureit,lu2019vilbert,karpathy2015alignment,lee2018stackedcrossattention}.
Many works employ standard architectures for processing images and text, such as CNNs-based models for image processing and recurrent networks for language.
Usually, in this scenario, the image embeddings are extracted from standard image classification networks, such as ResNet or VGG, by employing the network activations before the classification head. Usually, descriptions extracted from CNN networks trained on classification tasks are able to only capture global summarized features of the image, ignoring important localized details.

\begin{figure}[t]
    \centering
    \includegraphics[page=1, width=0.90\linewidth]{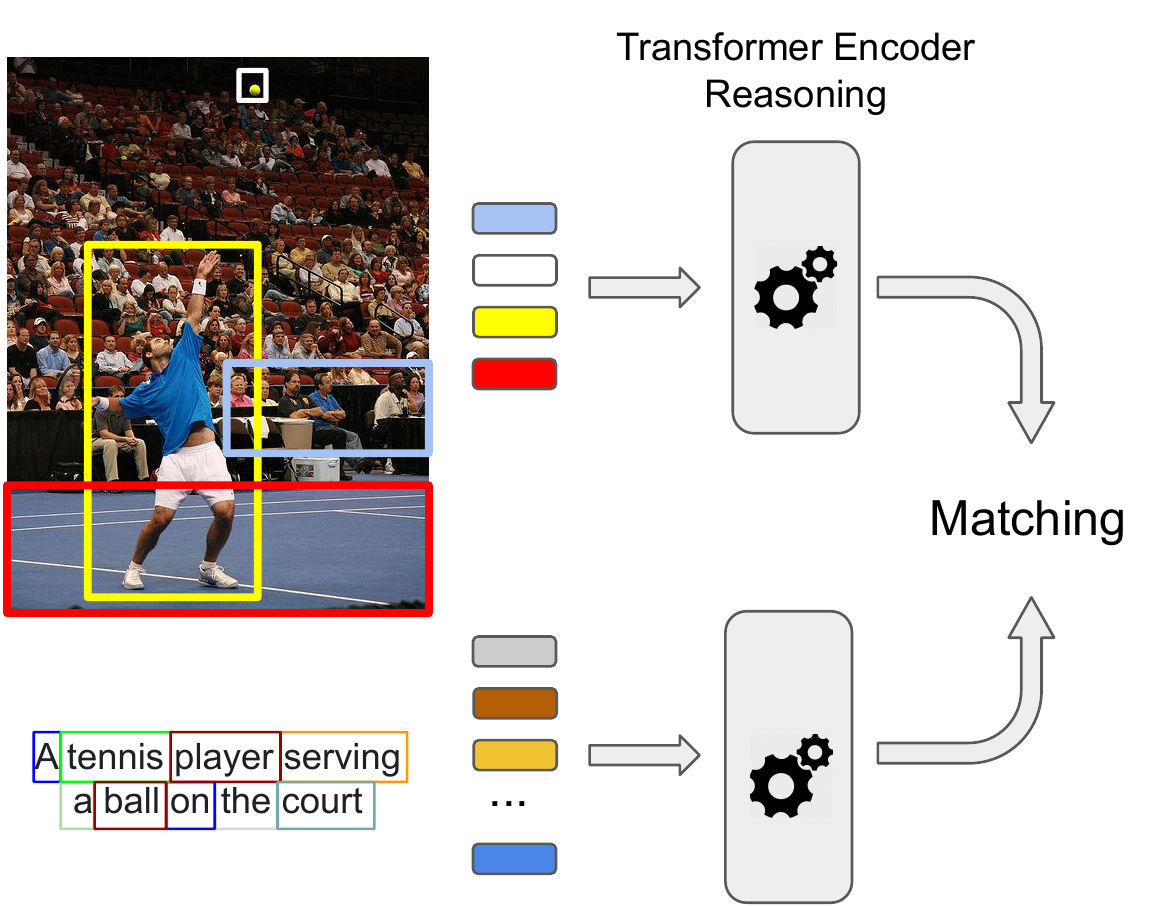}
  \caption{Overview of the presented architecture. Image and text are seen respectively as sets of image regions and sequences of words, and they are processed using a transformer-based reasoning engine.}
  \label{fig:simple_architecture} 
\end{figure}

Although these networks demonstrated noticeable performances in the image-text matching task, they are not able to infer what an \textit{object} really is. The \textit{objectness} prior is an important feature of the perception system that helps filtering out irrelevant zones in the images while focusing the attention on entities of interest. As far as the matching problem is concerned, finding entities of interest inside the image helps in creating a representation that has a level of abstraction comparable with the related text. In fact, a visual object present in an image, such as a dog, can be matched in an almost one-to-one relationship with the nouns \textit{dog}, or \textit{animal} present in the corresponding image caption. Furthermore, the \textit{objectness} prior is the first step towards higher-level abstraction tasks such as reasoning about inter-object relationships.

We are to tackle this important problem with the goal of finding compact cross-modal descriptions of images and texts which can incorporate detailed relational insights of the scene. Compact and informative descriptions are required in the context of large scale retrieval systems, where image and text embeddings can be compared and indexed using a simple similarity function (e.g., cosine similarity) defined on a common embedding space.

Some works have recently tackled the matching problem using a relational approach, trying to reason on substructures of images and texts (regions and words respectively) using attention and self-attention mechanisms \cite{qi2020imagebert,lu2019vilbert,lee2018stackedcrossattention}, or graph networks \cite{li2019}.

In particular \cite{qi2020imagebert,lu2019vilbert,karpathy2015alignment} try to learn a scoring function $s = \phi(I, C)$ measuring the affinity between an image and a caption, where $I$ is an image, $C$ is the caption and $s$ is a normalized score in the range $[0, 1]$.
The problem with this approach is that it is not possible to extract compact features describing images and texts separately. In this setup, if we want to retrieve images related to a given query text, we have to compute all the similarities through the $\phi$ function, and then sort the resulting scores in descending order. This is unfeasible if we want to retrieve images from a large database in few milliseconds.

For this reason, we propose the Transformer Encoder Reasoning Network (TERN), a transformer-based architecture able to map images and texts into the same common space while preserving important relational aspects of both modalities. In doing so, we avoid cross-talking between the two pipelines, so that it remains possible to separately forward the visual and the language pipeline to obtain compact image/caption descriptors.

The general transformer architecture \cite{vaswani2017transformer} was introduced to process sequential data, like natural languages. However, the encoder part of the transformer has no sequential prior hard-coded in its architecture. Therefore, it is a good candidate for processing also image regions: with the very desirable self-attention mechanism it incorporates, the transformer encoder can be employed to link together different image regions, effectively constructing a powerful visual reasoning pipeline. 

Concerning the evaluation of the proposed matching procedure in an information retrieval setup, the Recall@K metric is usually employed, where typically $K = \{1, 5, 10\}$.
However, in common search engines where the user is searching for related images and not necessarily exact matches, the Recall@K evaluation is too rigid \highlight{and unable to capture high-level semantic similarities between the retrieval results.}

For this reason, we propose to measure the retrieval abilities of the system through a discounted cumulative gain metric with relevance computed exploiting caption similarities \highlight{working at a high conceptual level}, proceeding in a similar way to \cite{carrara2018pictureit}.

% For text we can use a similar architecture, considering words as the counterparts of the image regions. This is basically what the BERT architecture \cite{devlin2019bert} was designed for. BERT uses the transformer encoder to learn inter-word dependencies, and BERT embeddings recently became a new standard for obtaining context-aware word features.

%In order to enforce high-level scene description and relational understanding in the learned common space, we also try to impose the reconstruction of the sentence and the region features given the vectors mapped in the common space.

The contributions of this paper are:
\begin{itemize}
    \item We introduce the Transformer Encoder Reasoning Network (TERN), a transformer-based architecture able to map both visual and textual modalities into the same common space, preserving the relational content of both modalities. The learned representations can be used for efficient and scalable multi-modal retrieval.
    \item We introduce a novel evaluation metric able to capture non-exact search results, by weighting different results through a relevance measure computed on the caption similarities.
    \item We show that our architecture reaches state-of-the-art performances with respect to other architectures on the introduced metric, for the image retrieval task.
    %\item We show that by enforcing the reconstruction of region features and sentences we are able to regularize the training, obtaining best performances on the retrieval task.
\end{itemize}

\section{Related Work}
In this section, we review some of the previous works related to image-text matching and high-level relational reasoning. Also, we briefly summarize the evaluation metrics available in the literature for the image-caption retrieval task.

\subsection*{Image-Text matching}
Image-text matching is often cast to the problem of inferring a similarity score among an image and a sentence. Usually, one of the common approaches for computing this cross-domain similarity is to project images and texts into a common representation space on which some kind of similarity measure can be defined (e.g.: cosine or dot-product similarities).
%Images and sentences are preprocessed by specialized architectures before being merged together at some point in the pipeline.

Concerning image processing, the standard approach consists in using Convolutional Neural Networks (CNNs), usually pre-trained on image classification tasks. In particular, some works \cite{KleinLSW15fishervectors,VendrovKFU15,LinP16,HuangWW17,EisenschtatW17} used VGGs, others \cite{LiuGBL17,vsepp2018faghri,GuCJN018,Huang2018} used ResNets. 
The problem with these kinds of CNNs is that they usually are able to extract extremely summarized and global descriptions of images. Therefore, a lot of useful fine-grained information needed to reconstruct inter-object relationships for precise image-text alignment is permanently lost.
\highlight{To overcome this problem, the authors in \cite{sun2015automatic} worked on a fine-grained level, defining the image-sentence matching score on the basis of the similarities among the common abstract concepts found in the two modalities.}

Recent works, instead, exploited the availability of precomputed region-level features extracted from state-of-the-art object detectors. In particular, following the work by \cite{AndersonHBTJGZ17}, the authors in \cite{li2019,lee2018stackedcrossattention} used bottom-up features extracted from Faster-RCNN. The bottom-up attention mechanism resembles the attentive mechanism present in the human visual system, and it is an important feature for filtering out unimportant information. This lays the foundations for a more precise and lightweight reasoning mechanism, downstream of the bottom-up perception module, which should carefully process the resulting image regions to obtain an expressive representation of the overall scene.

Concerning sentence processing, many works \cite{karpathy2015alignment,vsepp2018faghri,li2019,lee2018stackedcrossattention,Huang2018} employed GRU or LSTM recurrent networks to process natural language.

Recently, the transformer architecture \cite{vaswani2017transformer} achieved state-of-the-art results in many natural language processing tasks, such as next sentence prediction or sentence classification. In particular, the BERT embeddings \cite{devlin2019bert} emphasized the power of the attention mechanism to produce accurate context-aware word descriptions.

Given the enormous flexibility of the transformer encoder architecture, some works \cite{lu2019vilbert,qi2020imagebert} tried to apply the attention mechanism of the transformer encoder architecture to process visual inputs and natural language together. 
The main idea behind visual processing using the transformer encoder is to leverage its self-attention mechanism to link together different image regions to catch important inter-object relationships. %This is possible since this model is perfectly agnostic on the nature of the vectors given as input, and has no built-in sequential priors. 
\highlight{However, the systems proposed in this direction are not able to extract separate visual and textual features for use in similarity search applications.}

%These latest works were able to achieve state-of-the-art results in caption/image retrieval. However, they cannot separately produce image and caption embeddings; this is a mandatory requirement to produce features that are actually usable in real-world search engines. 
%They model a function $s = \phi(I, C)$ that measures the affinity between an image and a caption, where $I$ is an image, $C$ is the caption and $s$ is a normalized score in the range $[0, 1]$. Following this path, an exhaustive sequential search is needed to rank all the images given a query caption or vice-versa.

%Instead, we are interested in employing two different mapping functions, $I_v = \psi_v(I)$ and $I_c = \psi_t(C)$ which separately project the two modalities into the same common space, without preconditioning one of the modalities to the other.

The authors in \cite{li2019} were able to achieve very good results in caption/image retrieval learning separate visual and textual features. They introduced a visual reasoning pipeline built of a Graph Convolution Networks (GCNs) and a GRU to sequentially reason on the different image regions. Furthermore, they impose a sentence reconstruction loss to regularize the training process.
Differently from their work, we leverage on the reasoning power of the transformer encoder, both for the visual and linguistic pipelines. %Moreover, we try to impose a permutation-invariant region reconstruction loss to further regularize the training process.

\subsection*{High-level reasoning}
Another branch of research is focused on the study of relational reasoning models for high-level understanding. The authors in \cite{santoro2017rn} proposed an architecture that separates perception from reasoning. They tackled the problem of Visual Question Answering by introducing a particular layer called Relation Network (RN), which is specialized in comparing pairs of objects. Object representations are learned using a four-layer CNN, and the question embedding is generated through an LSTM. Recently, the authors in \cite{messina2019avfrn,DBLP:messina2019cbir} extended the RN for producing compact features for relation-aware image retrieval. However, they did not explore the multi-modal retrieval setup.

Other solutions try to stick more to a symbolic-like way of reasoning. In particular, some works \cite{learning_to_reasoning_end_to_end,inferring_and_executing_programs} introduced compositional approaches able to explicitly model the reasoning process by dynamically building a reasoning graph that states which operations must be carried out and in which order to obtain the right answer.

Recent works employed Graph Convolution Networks (GCNs) to reason about the interconnections between concepts. In particular, the works by \cite{YaoPLM18,YangTZC19,LiJ19} used GCNs to reason on the image regions for image captioning, while others \cite{YangLLBP18graphrcnn,LiOZSZW18} used GCN with attention mechanisms to produce the scene graph from plain images.

\subsection*{Retrieval evaluation metrics}
All the works involved with image-caption matching evaluate their results by measuring how good the system is at retrieving relevant images given a query caption (image-retrieval) and vice-versa (caption-retrieval). In other words, they evaluate their proposed models using a retrieval setup.

Usually the \textit{Recall@}K metric is used \cite{vsepp2018faghri,li2019,qi2020imagebert,lu2019vilbert,lee2019}, where typically $K = \{1, 5, 10\}$.
On the other hand, the authors in \cite{carrara2018pictureit} introduced a novel metric able to capture non-exact results by weighting the ranked documents using a caption-based similarity measure.
We embrace their idea, and we extend it bringing to life a powerful evaluation metric \highlight{able to capture high-level semantic aspects. Furthermore,} relaxing the constraints of exact-match similarity search is an important step towards an effective evaluation of modern search engines.

\section{Review of Transformer Encoders (TEs)}
Our proposed architecture is based on the well established Transformer Encoder (TE) architecture, which relies heavily on the concept of self-attention.
The basic attention mechanism, as described in \cite{vaswani2017transformer}, is built upon three quantities: queries, keys, and values.
The attention mechanism maps a query and a set of key-value pairs to an output, where the query, keys, values, and output are all vectors. The output is computed as a weighted sum of the values, where the weight assigned to each value is computed using a softmax activation function over the inner product of the query with the corresponding key.
More formally,
\begin{equation}
    \text{Att}(Q, K, V) = \text{softmax}\left ( \frac{QK^T}{\sqrt{d_k}} \right ) V,
\end{equation}
where $Q \in \mathbb{R}^{t \times d_k}, K \in \mathbb{R}^{s \times d_k}$ and $V \in \mathbb{R}^{s \times d_v}$; $s$ is the input sequence length and $t$ is the length of the conditioning sequence that drives the attention. The factor $\sqrt{d_k}$ is used to mitigate the vanishing gradient problem of the softmax function in case the inner product assumes too large values.

The self-attention derives trivially from the general attention mechanism when either $V$, $K$, and $Q$ are computed from the same input set, i.e., when the set that we use to drive the attention is the same as the input set. In this case, in fact, $t = s$ and the scalar product $QK^T \in \mathbb{R}^{s \times s}$ is a square matrix that encodes the affinity that each element of the set has with all the others elements of the same set.

In the self-attention case, $Q$, $K$, and $V$ are computed by linearly projecting the same input embeddings using three different matrices $W^Q\in \mathbb{R}^{d_k \times d_i}, W^K\in \mathbb{R}^{d_k \times d_i}$ and $W^V\in \mathbb{R}^{d_v \times d_i}$, where $d_i$ is the dimensionality of the input embeddings.

\begin{figure}[t]
    \centering
    \includegraphics[page=4, width=0.90\linewidth]{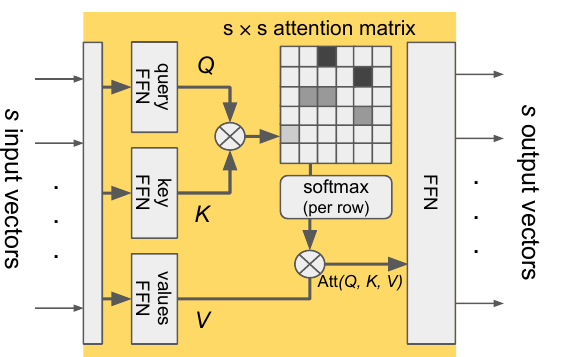}
  \caption{\highlight{Simplified view of a transformer encoder layer. Add\&Norm skip connections present in the original architecture are not shown here for ease of viewing.}}
  \label{fig:TE} 
\end{figure}

%The self-attention is the core building block of the transformer encoder (TE) architecture. 
Then, a simple feedforward layer on the $\text{Att}(Q, K, V)$ vectors, with a ReLU activation function, further processes the vectors produced by the self-attention mechanism. %This simple feedforward layer casts in output a set of features having the same dimensionality of the input set. 
A schematic view of a transformer encoder layer is shown in Figure \ref{fig:TE}.

We argue that the transformer encoder self-attention mechanism is able to drive a simple but powerful reasoning mechanism able to spot hidden links between the vector entities given in input to the encoder, whatever nature they have. Also, the encoder is designed in a way that multiple instances of the same architecture could be stacked in sequence, producing a deeper reasoning pipeline.

\section{Visual-Textual Reasoning using Transformer Encoders}
Our work relies almost entirely on the TE architecture, both for the visual and the textual data pipelines. 
%The TE achieved state-of-the-art results on many natural language processing tasks and it has been recently used to process regions extracted from images. The main reason for using the TE lies in its self-attentive module, which is able to create meaningful connections between related concepts, effectively creating a simple but powerful reasoning engine.

The TE takes as input sequences or sets of entities, and it can reason upon these entities disregarding their intrinsic nature.
In particular, we consider the salient regions in an image as visual entities, and the words present in the caption as language entities.

More formally, the input to our reasoning pipeline is a set $I = \{r_0, r_1, ... r_n\}$ of $n$ image regions representing an image $I$ and a sequence $C = \{w_0, w_1, ... w_m\}$ of $m$ words representing the corresponding caption $C$.
Following, we will describe the methodology we adopted to extract $r_i$ from images and $w_j$ from captions.

\begin{figure*}[ht]
    \centering
    \includegraphics[page=2, width=0.70\linewidth]{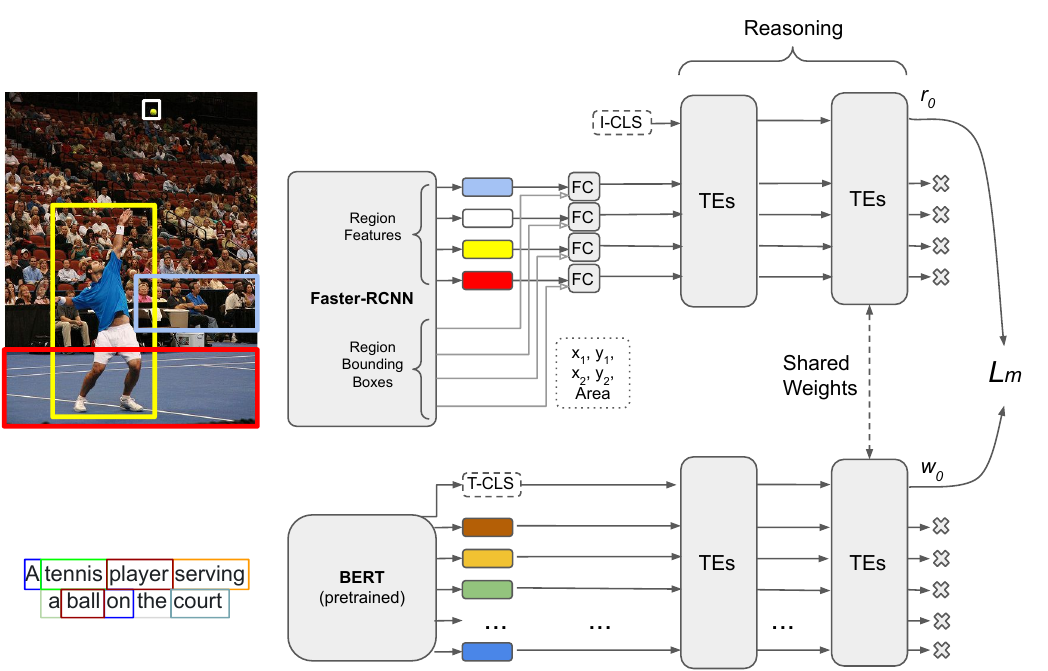}
  \caption{The proposed TERN architecture. TE stands for Transformer Encoder and its architecture is explained in detail in \cite{vaswani2017transformer}. Region and words are extracted through a bottom-up attention model based on Faster-RCNN and BERT respectively. BERT already employs positional encoding for representing the sequential nature of words, therefore this step is not reported in the figure. Concerning regions, the extracted bottom-up features are conditioned with the information related to the geometry of the bounding-boxes. This is done through a simple fully connected stack in the early visual pipeline, before the reasoning steps. $L_m$ is the matching loss, defined as in \cite{vsepp2018faghri}. The final weight sharing between TE modules guarantees consistent processing of the high-level concepts.}.
  \label{fig:detailed_architecture} 
\end{figure*}

\subsection*{Region and Word Features}
$I$ and $C$ descriptions come from state-of-the-art visual and textual pre-trained networks, Faster-RCNN with Bottom-Up attention, and BERT respectively.

Faster-RCNN \cite{RenHGS15fasterrcnn} is a state-of-the-art object detector. It has been used in many downstream tasks requiring salient object regions extracted from images. 
%Therefore, Faster-RCNN is one of the main architectures implementing the human-like visual perception.
The authors in \cite{Anderson2018bottomup} introduced bottom-up visual features by training Faster-RCNN with a Resnet-101 backbone on the Visual Genome dataset \cite{Krishna2016VisualGenome}. Using these features, they were able to reach remarkable results on the two downstream tasks of image captioning and visual question answering.
Therefore, in our work we employ the bottom-up features extracted from every image as image description $I = \{r_0, r_1, ... r_n\}$.

Concerning text processing, we used BERT \cite{devlin2019bert} for extracting word embeddings. BERT already uses a multi-layer transformer encoder to process words in sentences and capture their functional relationships through the same powerful self-attention mechanism. BERT embeddings are trained on some general natural language processing tasks such as sentence prediction or sentence classification and demonstrated state-of-the-art results in many downstream natural language tasks.
BERT embeddings, unlike word2vec \cite{Mikolov2013word2vec}, capture the context in which each word appears. Therefore, every word embedding carries information about the surrounding context, that could be different from caption to caption.
%Since the transformer encoder architecture does not embed any sequential prior in its architecture, words are given a sequential order by mixing some positional information into the learned input embeddings. For this reason, the authors in \cite{vaswani2017transformer} add sine and cosine functions of different frequencies to the input embeddings. This is a simple but effective way to transform a set into a sequence.

\subsection*{Transformer Encoder Reasoning Network (TERN)}

Our reasoning engine is built using a stack of transformer encoder layers; the overall architecture is shown in Figure \ref{fig:detailed_architecture}. %The same reasoning architecture is applied to both the textual and visual pipelines.

The reasoning module continuously operates on sets and sequences of $n$ and $m$ objects respectively for images and captions. 
The objective is to produce a compact representation of the $n$ processed regions and of the $m$ processed words suitable for the downstream task of image-text matching in a common space with fixed dimensionality.
One of the easiest ways to proceed is to pool the elements of the set/sequence using symmetric functions like \textit{sum} or \textit{avg}, or, like \cite{li2019}, growing a meaningful aggregated representation inside the hidden state of a recurrent network (GRU or LSTM).

Our method, instead, follows the approach by BERT \cite{devlin2019bert}: we reserve a special token both at the beginning of the regions set and of the words sequence (I-CLS and T-CLS) devoted to carrying global information along the two pipelines. 
For this reason, we effectively expand the number of image regions to $n+1$ and the number of words to $m+1$, with $r_0$ and $w_0$ reserved for this purpose. 
Initially, $w_0$ is set to the T-CLS BERT token, while $r_0$, i.e., I-CLS, is a zero vector.
At every reasoning step, this information is updated attentively by the self-attention mechanism of the TEs.
In the end, our final image and caption features will be $r_0$ and $w_0$ in output from the last transformer encoder layer.
In the last layers of the TERN architecture, the abstracted representations of the visual and textual pipelines should be comparable. To enforce this constraint, we share the weights of the last layers of the TEs before computing the matching loss $L_m$ on the common space.

If we use only bottom-up features without any spatially related information, the visual reasoning engine is not able to reason about spatial relationships. This is a fairly important aspect to capture since lot of textual descriptions contain spatial indications (e.g. \textit{on top of} or \textit{above}). 
In order to include spatial awareness also in the visual reasoning process, we condition the early visual pipeline with the bounding-boxes coordinates. To this aim, we compute the normalized coordinates and the normalized area for each region, as follows:
\begin{equation}
    c = \left\{\frac{x_1}{W}, \frac{y_1}{W}, \frac{x_2}{H}, \frac{y_2}{H}, \frac{(x_2 - x_1)(y_2 - y_1)}{WH}\right\}.
\end{equation}

Then, we concatenate $c$ with the original bottom-up feature. In the end, we forward this information through a simple Linear-ReLU-Linear stack (sharing weights among all the $n$ regions) to obtain the final spatial-aware bottom-up feature.

\subsection*{Learning}
In order to match images and captions in the same common space, we use a hinge-based triplet ranking loss, focusing the attention on hard negatives, as in \cite{vsepp2018faghri,li2019}.
Therefore, we used the following loss function:
\begin{equation}
\begin{split}
    L_m(i, c) &= \max_{{c}'} [\alpha + S(i, {c}') - S(i, c)]_+ + \\
    &\mathrel{\phantom{=}} \max_{{i}'} [\alpha + S({i}', c) - S(i, c)]_+,
\end{split}
\end{equation}
where $[x]_+ \equiv max(0, x)$. The hard negatives ${i}'$ and ${c}'$ are computed as follows: 
\begin{equation}
\begin{split}
    {i}' = \text{arg} \max_{j \neq i} S(j, c) \\
    {c}' = \text{arg} \max_{d \neq c} S(i, d),
\end{split}
\end{equation}
where $(i, c)$ is a positive pair.
$S(i,j)$ is the similarity function between image and caption features. We used the standard cosine similarity as $S(\cdot, \cdot)$.
As in \cite{vsepp2018faghri}, the hard negatives are sampled from the mini-batch and not globally, for performance reasons.

\section{Evaluation Metric for Non-Exact Matching}
Many works measure the retrieval abilities of their visual-linguistic matching system by employing the well known \textit{Recall@}K metric. The Recall@K measures the percentage of queries able to retrieve the correct item among the first k results.

However, in common search engines where the user is searching for related images/captions and not necessarily exact matches, the Recall@K evaluation \highlight{often proves to be} too rigid, especially when $K$ is small. In fact, in the scenarios where $K = \{1, 5, 10\}$, we are measuring the ability of the system to retrieve exact results at the top of the ranked list of images/captions. Doing so, we are completely ignoring other relevant but not exact-matching elements retrieved in the first positions. These elements still contribute to a good user experience in the context of search engines. 
The Recall@K metric is fully unable to capture this simple yet important aspect. \highlight{Furthermore, Recall@K is unable to capture high-level semantics of sentences and images during the retrieval evaluation phase}.

For this reason, inspired by the evaluation method presented in \cite{carrara2018pictureit}, we employed a common metric often used in information retrieval applications, the Normalized Discounted Cumulative Gain (NDCG). 
The NDCG is able to evaluate the quality of the ranking produced by a certain query by looking at the first $p$ position of the ranked elements list. 
The premise of NDCG is that highly relevant items appearing lower in a search result list should be penalized as the graded relevance value is reduced proportionally to the position of the result.

The non-normalized DCG until position $p$ is defined as follows:
\begin{equation}
    \text{DCG}_{p} = \sum _{i=1}^{p}{\frac {\text{rel}_{i}}{\log _{2}(i+1)}},
\end{equation}
where $\text{rel}_i$ is a positive number encoding the affinity that the $i$-th element of the retrieved list has with the query element. The DCG is agnostic upon how the relevance is computed.
The $\text{NDCG}_p$ is computed by normalizing the $\text{DCG}_p$ with respect to the Ideal Discounted Cumulative Gain (IDCG), that is defined as the DCG of the list obtained by sorting all its elements by descending relevance:
\begin{equation}
    \text{NDCG}_{p} = \frac{\text{DCG}_p}{\text{IDCG}_p}.
\end{equation}

$\text{IDCG}_p$ is the best possible ranking. Thanks to this normalization, $\text{NDCG}_p$ acquires values in the range $[0, 1]$.

\subsection*{Computing $\text{rel}_i$ values}
We concentrate our attention on image-retrieval, given that is the most common scenario in real-world search engines. Therefore, in our work, we consider a caption as a query, while the retrieved elements are images.
%In a cross-modal retrieval setup, there are two possible retrieval scenarios:
%\begin{itemize}
%    \item The query is an image and the retrieved elements are captions (sentence-retrieval)
%    \item The query is a caption and the retrieved elements are images (image-retrieval)
%\end{itemize}

Being a cross-modal retrieval setup, the relevance should be a value obtained from a function operating on an image $I_i$ and a caption $C_j$. In principle, it could be possible to use the $\phi(I_i, C_j)$ learned by methods like in \cite{lu2019vilbert,qi2020imagebert}. The problem is that $\phi$ is a complex neural network, and $I_i$, $C_j$ are drawn from a dataset of thousands of elements, in the best case. This means that constructing a $N_c \times N_i$ relevance matrix is computationally unfeasible, where $N_c$ is the number of total captions and $N_i$ is the total number of images in the dataset.

Usually, in the considered datasets, images come with a certain number of associated captions. Thus, instead of computing $\phi(I_i, C_j)$, we could think of computing $\tau(\bar{C}_i, C_j)$ instead, where $\bar{C}_i$ is the set of all captions associated with the image $I_i$.
With this simple expedient, we could efficiently compute quite large relevance matrices using similarity between captions, which in general are computationally much cheaper. 

As a result, for our image-retrieval objective we define $\text{rel}_i = \tau(\bar{C}_i, C_j)$, where $C_j$ is the query caption and $\bar{C}_i$ are the captions associated with the $i$-th retrieved image.
%\begin{itemize}
%    \item $rel_i = \tau(\bar{C}_i, C_j)$ in case of image retrieval, where $C_j$ is the query caption
%    \item $rel_i = \tau(\bar{C}_j, C_i)$ in case of caption retrieval, where $\bar{C}_j$ is the set of %captions associated to the query image $I_j$. 
%\end{itemize}

In our work, we use \texttt{ROUGE-L}\cite{lin-2004-rouge} and \texttt{SPICE}\cite{AndersonFJG16spice} as functions $\tau$ for computing captions similarities.
These two metrics capture different aspects of the sentences. In particular, \texttt{ROUGE-L} operates on the longest common sub-sequences, while \texttt{SPICE} exploits graphs associated with the syntactic parse trees, and has a certain degree of robustness against synonyms. In this way, \texttt{SPICE} is more sensitive to high-level features of the text and semantic dependencies between words and concepts rather than to pure syntactic constructions of the sentences.

\section{Experiments}
We train the Transformer Encoder Reasoning Network and we measure its performance on the MS-COCO \cite{LinMBHPRDZ14coco} dataset, by measuring the effectiveness of our approach on the image retrieval task. We compare our results against state-of-the-art approaches on the same dataset, using the introduced metric.

The MS-COCO dataset comes with a total of 123,287 images. Every image has associated a set of 5 human-written captions describing the image. 

We follow the splits introduced by \cite{karpathy2015alignment} and followed by the subsequent works in this field \cite{vsepp2018faghri,GuCJN018,li2019}. In particular, 113,287 images are reserved for training, 5000 for validating, and 5000 for testing.

At test time, results for both 5k and 1k image sets are reported. In the case of 1k images, the results are computed by performing 5-fold cross-validation on the 5k test split and averaging the outcomes.
%We computed caption-caption relevance for the NDCG metric using \texttt{ROUGE-L}\cite{lin-2004-rouge} and \texttt{SPICE}\cite{AndersonFJG16spice}. 
We set the NDCG parameter $p = 25$ as in \cite{carrara2018pictureit} in our experiments.

\subsection{Implementation Details}
We employ the BERT model pre-trained on the masked language task on English sentences, using the PyTorch implementation by HuggingFace \footnote{\url{https://github.com/huggingface/transformers}}. These pre-trained BERT embeddings are 768-D.
For the visual pipeline, we used the already available bottom-up features extracted on the MS-COCO dataset. They are freely available on GitHub \footnote{\url{https://github.com/peteanderson80/bottom-up-attention}} and they are 2048-D.
In the experiments we used the fixed-size descriptors, selecting for each image the features of the top 36 most confident detections. However, our pipeline can work with a variable-length set of regions for each image, by appropriately masking the attention weights in the TE layers.

Concerning the reasoning steps, we used a stack of 4 non-shared TE layers for visual reasoning. Instead, we found the best results when fine-tuning the pre-trained BERT, so we did not introduce any further non-shared TE layers for the language pipeline. 
We used 2 final TE layers with weights shared among the visual and textual pipelines. All the TEs feed-forward layers are 2048-dimensional and the dropout is set to 0.1.
Weight sharing in the last TE layers is possible if the input vectors from both visual and textual pipelines share the same number of dimensions. For this reason, before entering the last shared-weight TEs, both the visual and textual vectors are linearly projected to a 1024-D space, which is also the dimensionality of the final common space, as in \cite{vsepp2018faghri}.

We trained for 30 epochs using Adam optimizer with a learning rate of 0.00002.
The $\alpha$ parameter of the hinge-based triplet ranking loss is set to 0.2, as in \cite{vsepp2018faghri,li2019}.

We used a batch size of 90, instead of 128 as in previous works, due to hardware limitations. %, and we use a 1024-D common space, as in \cite{vsepp2018faghri}.

\begin{figure*}[ht]
    \centering
    \includegraphics[page=3, width=1\linewidth]{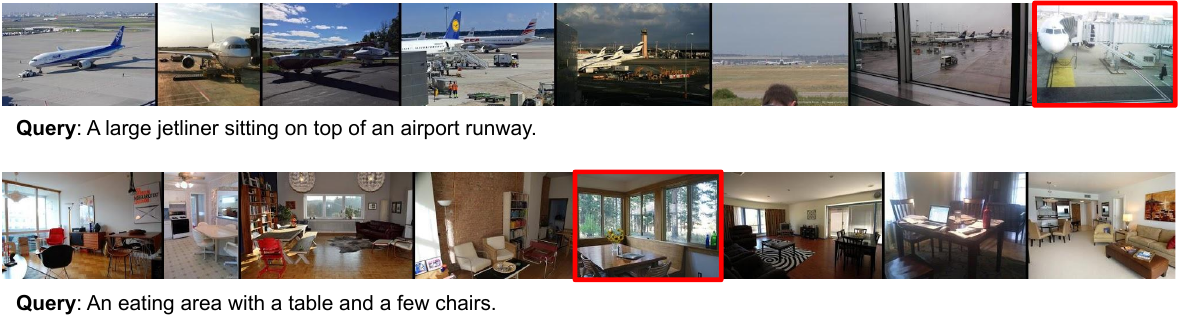}
  \caption{Example of image retrieval results for a couple of query captions. \highlight{The red marked images represent the MS-COCO ground truths, and they are not necessarily the best results in these scenarios. In fact, in the very first positions, we find non-matching yet relevant images.} These are common examples where NDCG really succeed over the Recall@K metric.}.
  \label{fig:examples} 
\end{figure*}

\renewcommand{\arraystretch}{1.2}
\newcolumntype{C}{>{\centering\arraybackslash}X}
\newcolumntype{R}{D{,}{\pm}{1.6}}
\newcolumntype{L}{>{\raggedright\arraybackslash}p{2cm}}
\begin{table}[t]
\caption{Image retrieval results on the MS-COCO dataset.}
\label{tab:results}
\begin{center}
\begin{tabular}{Lccccc}
\toprule
& \multicolumn{3}{c}{\textbf{Recall@K}} & \multicolumn{2}{c}{\textbf{NDCG}}\\
\cmidrule(lr){2-4} \cmidrule(lr){5-6}
\textbf{Model} & \multicolumn{1}{c}{K=1} & \multicolumn{1}{c}{K=5} & \multicolumn{1}{c}{K=10}
& \multicolumn{1}{c}{\texttt{ROUGE-L}} & \multicolumn{1}{c}{\texttt{SPICE}}\\
\midrule
\multicolumn{6}{c}{\textit{1K Test Set}} \\
\midrule
VSE0 \cite{vsepp2018faghri} & 43.7 & 79.4 & 89.7 & 0.702 & 0.616 \\
VSE++ \cite{vsepp2018faghri} & 52.0 & 84.3 & 92.0 & 0.712 & 0.617 \\
VSRN \cite{li2019} & 60.8 & 88.4 & 94.1 & 0.723 & 0.620 \\
TERN (Ours) & 51.9 & 85.6 & 93.6 & \textbf{0.725} & \textbf{0.653} \\
\midrule
\multicolumn{6}{c}{\textit{5K Test Set}} \\
\midrule
VSE0 \cite{vsepp2018faghri} & 22.0 & 50.2 & 64.2 & 0.633 & 0.549 \\
VSE++ \cite{vsepp2018faghri} & 30.3 & 59.4 & 72.4 & 0.656 & 0.577 \\
VSRN \cite{li2019} & 37.9 & 68.5 & 79.4 & \textbf{0.676} & 0.596 \\
TERN (Ours) & 28.7 & 59.7 & 72.7 & 0.665 & \textbf{0.600} \\
\bottomrule
\end{tabular}
\end{center}
\end{table}

\subsection{Results and Discussion}
We report the results obtained on the MS-COCO dataset on both 5k and 1k image test sets, and we compare them against the state-of-the-art on the image retrieval task.
For VSRN \cite{li2019} and VSE \cite{vsepp2018faghri} we used the original code and pre-trained models provided by the authors, updating the evaluation protocol by including the NDCG metric.

Concerning VSRN, in the original paper the results are given for an ensemble of two independently trained models.
In our case, we did not consider model ensembling. For this reason, we evaluated VSRN using the best snapshot among the two provided by the authors.

Results are reported in Table \ref{tab:results}. For the sake of completeness, we report also the values for the Recall@K metric.
Our TERN architecture can reach top performance on the NDCG metric with the \texttt{SPICE}-based relevance. Due to the high-level abstraction nature of the \texttt{SPICE} metric, this result confirms the ability of our system to understand complex patterns and abstract concepts both in the visual and textual inputs. \highlight{The NDCG metric with the \texttt{SPICE} relevance tries to measure the high-level perception and relational understanding abilities of the model directly on the downstream image-retrieval task.} We obtain the best gap on the 1K test set, where we improve the current state-of-the-art by 5.3\%.

Concerning the NDCG metric with the \texttt{ROUGE-L} computed relevance, our TERN architecture performs slightly worse than VSRN. Overall, the gap between VSRN and our TERN architecture is very subtle, confirming the ability of those architectures to be comparable when we focus on the syntactic and less abstract features of the language.

Despite VSRN performing better in terms of Recall@K, we demonstrated through the NDCG metric that our architecture is better at finding non-exact matching yet relevant elements in the top $p$ positions of the ranked images list. \highlight{The results on the new evaluation metric confirm the power of the TERN architecture to construct high-level visual-textual descriptions useful for similarity search in cross-modal environments}.

Figure \ref{fig:examples} shows an example of image retrieval using features computed through our TERN architecture. The reported examples show two typical situations in which the NDCG evaluation succeeds over the Recall@K. \highlight{The figure show how the ground truth images from the MS-COCO dataset, highlighted in red, are not necessarily the best retrieval results for the probed query sentences.}

\highlight{Having observed the qualitative and quantitative results, we argue that the Recall@K and NDCG metrics should be weighted appropriately when evaluating a cross-modal retrieval system. A system validated and tested only using Recall@K becomes strongly sensible to exact-matching elements only. This would result in a scenario where non-exact yet relevant matches are pulled away during the training phase, with no indicators to monitor this strongly unwanted scenario.
We think that these are very important considerations to keep in mind during the validation of cross-modal search engines.}

\section{Conclusions}
In this work, we addressed the problem of image-text matching in the context of efficient multi-modal information retrieval. We argued that many state-of-the-art methods do not extract compact features separately for images and text. This is a problem if we want to employ relationship-aware visual and textual features in the subsequent indexing stage for efficient and scalable cross-modal information retrieval.
To this aim, we developed a relationship-aware architecture based on the Transformer Encoder (TE) architecture to reason about the spatial and abstract relationships between elements in the image and the text separately. The final weight sharing between TE modules guarantees consistent processing of the high-level concepts.

In the vision of employing this architecture for efficient multi-modal information retrieval in real-world search engines, we measured our results using an NDCG metric assessing possibly non-exact but relevant search results. The relevance among images and captions has been evaluated by employing similarity measures defined over captions, \texttt{ROUGE-L} and \texttt{SPICE} respectively. We demonstrated that our relation-aware approach for reasoning and matching visual and textual concepts achieved state-of-the-art results with respect to current multi-modal matching architectures on the proposed retrieval metric, for the task of image retrieval.

In the near future, we manage to enforce some reconstruction constraints for better shaping the common space, like reconstructing the sentences from the visual features, as in \cite{li2019}, or recovering the image regions from the captions. Also, major interest should be given to the optimization objective. In particular, it would be interesting to attenuate the very aggressive behavior of the hinge-based triplet ranking loss for better appreciating non-exact matches at training time.


\section*{Acknowledgment}
This work was partially funded by:
AI4Media - A European Excellence Centre for Media, Society and Democracy (EC, H2020 n. 951911);
AI4EU project (EC, H2020, n. 825619);
AI4ChSites, CNR4C program (Tuscany POR FSE 2014-2020 CUP B15J19001040004).




\bibliographystyle{IEEEtran}
\bibliography{biblio.bib}
%\begin{thebibliography}{1}

%\end{thebibliography}

\end{document}